\documentclass{article}
\usepackage{ijcai25}

\usepackage{times}
\usepackage{soul}
\usepackage{url}
\usepackage[hidelinks]{hyperref}
\usepackage[utf8]{inputenc}
\usepackage[small]{caption}
\usepackage{graphicx}
\usepackage{amsmath}
\usepackage{amsthm}
\usepackage{booktabs}
\usepackage{algorithm}
\usepackage{algorithmic}
\usepackage[switch]{lineno}
\usepackage{xcolor, multirow}

\title{Fair-MoE: Fairness-Oriented Mixture of Experts in Vision-Language Models\\}

\author{
Peiran Wang$^1$
\and
Linjie Tong$^1$\and
Jiaxiang Liu$^2$\and
Zuozhu Liu$^2$\\
\affiliations
$^1$University of Illinois Urbana-Champaign\
$^2$Zhejiang University\\
\emails
\{peiranw3, linjiet2\}@illinois.edu,
\{jiaxiang.21, zuozhuliu\}@intl.zju.edu.cn
}
\definecolor{darkgreen}{rgb}{0.0, 0.5, 0.0} % 定义暗绿色

\begin{document}

\maketitle

\begin{abstract}
    Fairness is a fundamental principle in medical ethics. Vision Language Models (VLMs) have shown significant potential in the medical field due to their ability to leverage both visual and linguistic contexts, reducing the need for large datasets and enabling the performance of complex tasks. However, the exploration of fairness within VLM applications remains limited. Applying VLMs without a comprehensive analysis of fairness could lead to concerns about equal treatment opportunities and diminish public trust in medical deep learning models. To build trust in medical VLMs, we propose Fair-MoE, a model specifically designed to ensure both fairness and effectiveness. Fair-MoE comprises two key components: \textit{the Fairness-Oriented Mixture of Experts (FO-MoE)} and \textit{the Fairness-Oriented Loss (FOL)}. FO-MoE is designed to leverage the expertise of various specialists to filter out biased patch embeddings and use an ensemble approach to extract more equitable information relevant to specific tasks. FOL is a novel fairness-oriented loss function that not only minimizes the distances between different attributes but also optimizes the differences in the dispersion of various attributes' distributions. Extended experiments demonstrate the effectiveness and fairness of Fair-MoE. Tested on the Harvard-FairVLMed dataset, Fair-MoE showed improvements in both fairness and accuracy across all four attributes. Code will be publicly available.
\end{abstract}

\section{Introduction}
Fairness is a fundamental principle of medical ethics \cite{varkey2021principles,wang2023multidisciplinary,giovanola2023beyond,pratt2020justice}, emphasizing the need for equitable decision-making processes in diagnosis. It requires that diagnostic systems avoid systematically disadvantaging specific groups–such as black and white, male and female–based on inherent or acquired characteristics in a protected attribute like race or gender\cite{parraga2023fairness,luo2024fairclip,mehrabi2021survey}. A prominent example of fairness issues in the medical field is the disparity in diagnostic accuracy\cite{ferrante2022addressing,xu2024addressing,stanley2022fairness}, where certain groups experience significantly lower accuracy compared to others within the same diagnostic framework. Such discrepancies can result in worse patient outcomes for these groups, further exacerbating existing healthcare disparities. 

However, such fairness issues are concealed in some deep learning models, which are increasingly used in diagnostic tools. These models may inadvertently perpetuate biases due to imbalances in training datasets or insufficient representation of specific demographic groups. \cite{luo2024fairclip,glocker2023algorithmic,khan2023fair,sikstrom2022conceptualising}.
This not only exacerbates healthcare disparities but also undermines public trust in medical systems. Therefore, developing a fair and unbiased deep learning model for diagnostics is both critical and essential, especially in demographically diverse societies where achieving healthcare equality is a top priority.  

Meanwhile, Vision Language Models (VLMs) are widely applied in medical image analysis 
\cite{liu2023parameter,liu2023chatgpt,gai2024medthink,huang2023visual,qin2022medical,liu2024medcot,liu2024vpl,liu2025kpl} and represent a growing trend.
Natural pairings of medical images and reports, which require no additional annotation, can be used to fine-tune VLMs directly, eliminating the time-consuming process of data annotation.
Additionally, VLMs are designed to process and integrate visual and textual data simultaneously, enabling the handling of complex tasks by leveraging both visual and linguistic contexts. By analyzing medical images alongside associated textual information, such as doctors' notes, VLMs can achieve a more nuanced understanding of the inputs and make more accurate decisions. However, like other AI models, VLMs are not immune to fairness problems. Biases in datasets, such as those related to race, gender, and ethnicity, can lead to disparities in performance across demographic groups. Given the significant potential for VLM applications in the medical field, exploring and ensuring fairness within these models is both essential and valuable.

Despite the importance of exploring fairness in VLMs, tackling and resolving biases in these models has been challenging. The inclusion of text modality in VLMs introduces additional complexity compared to vision-only models (VMs)\cite{luo2024fairclip}, as biases in textural data can compound or interact with biases in visual data, exacerbating disparities\cite{guan2024eclb,tong2023tsnet,liu2023deep,yang2021infrared}.  Furthermore, while fairness research in VMs has gained traction–such as studies revealing biases in natural image domains\cite{qiang2023fairness,rohdr} and exposing inequities in VMs processing X-ray images in medical contexts\cite{glocker2023algorithmic,khan2023fair}–there remains a critical lack of open dataset focused on bias evaluation and mitigation in VLMs\cite{kelly2024visiongpt,kelly2024visiongpt1,luo2022biogpt,wang2022medclip,xu2023elixr,yang2024worldgpt}. Until recently, the release of the Harvard-FairVLMed multi-modal image-text dataset\cite{luo2024fairclip}, designed to explore fairness in medical VLMs, offers an unprecedented opportunity to address these challenges. 

Harvard-FairVLMed \cite{luo2024fairclip} contains 10000 image-text pairs about glaucoma. In addition to the ground truth of glaucoma, FairVLMed provides four protected attributes: gender, race, ethnicity, and language. 
Based on the Harvard-FairVLMed dataset, FairCLIP \cite{luo2024fairclip}, a benchmark for fair VLM, has been proposed. However, while FairCLIP relies on minimizing Sinkhorn distance \cite{peyre2019computational} to enhance fairness, it retains CLIP’s original architecture without specific adaptations for fairness considerations. This limitation restricts FairCLIP’s ability to effectively learn fair information from data.
% Thus, despite achieving SOTA performance in a trade-off between accuracy and fairness, FairCLIP fails to improve both fairness and accuracy. 
% However, in medical realm, both effectiveness and fairness are vital\cite{9308226}, and improving one of them without sacrificing another one is a more favorable situation.  
Due to the limitations of FairCLIP and the pressing need for a fair and effective VLM in the medical domain, there is a necessity for a new VLM. This model should enhance both accuracy and fairness by focusing on extracting task-relevant information while disregarding biased information that is not directly pertinent to the task. This objective demands advanced learning capabilities, which can be achieved using a Mixture of Experts-based model (MoE).
MoE demonstrates that a complex task can be decomposed into several subtasks, and that combining several experts, typically Multilayer Perceptrons (MLPs), through a gating mechanism, can achieve a competitive ability to learn and solve complex tasks \cite{6797059}. 

Recently, numerous variants of MoEs have been proposed \cite{riquelme2021scaling,lou2021cross,fedus2022switch,zoph2022st}, significantly enhancing the model's learning capabilities and yielding better results. 
Despite MoE's strong learning capabilities, efforts to achieve fairness in VLMs using MoE are limited. This can serve as a promising framework for exploring fairness.
Current research on developing MoE-based fair algorithms mainly focuses on classical machine learning problems where features, such as from images or texts, are predefined and the model primarily performs classification tasks \cite{germino2024fairmoe,sharma2022feamoe}. There is a lack of development in MoE architectures specifically tailored for fair medical vision language models. This presents a significant opportunity for advancing both the development of MoEs and their application in creating fair and effective VLMs for medical diagnostics. 

\begin{figure}[t]
    \centerline{\includegraphics[width=1.0\linewidth]{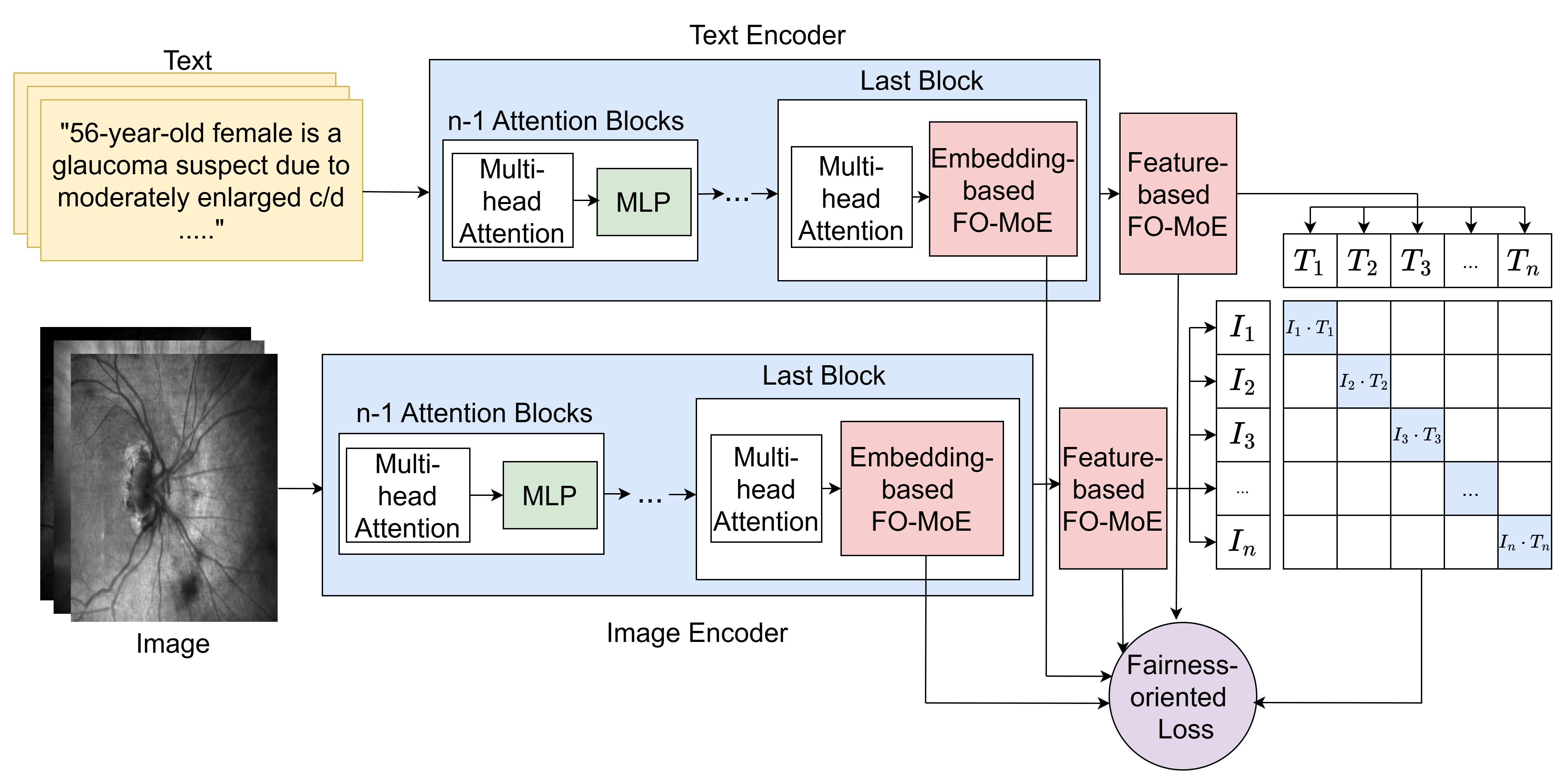}}
    \caption{An illustration of Fair-MoE architecture. MoE-based architecture with FO-MoE enables model to extract fair information. Model is trained through contrastive learning with novel fair loss FOL added. }
    \label{fig1}
\end{figure}

To address the aforementioned issues and pursue more equitable and accurate VLMs, we propose the first MoE-based model for fair medical vision language applications: the Fair Medical Vision Language Mixture of Experts (Fair-MoE) Model. This model comprises two advanced modules: \textit{Fairness-Oriented Mixture of Expert (FO-MoE)} and \textit{Fairness-Oriented Loss (FOL)}. FO-MoE is the first MoE that designs for fair VLM. It employs expert capacity to filter out bias patch embedding, thereby enhancing the model’s learning ability to extract more fair task-relevant information while reducing the likelihood of extracting biased, task-irrelevant information. Unlike other fair losses that focus on distance between different protected attributes, FOL is a novel fair load balance loss function that takes both distance and dispersion between protected attributes into account. In this way, it can not only guarantee fairness but also enhance the learning ability of the MoE. Fig. \ref{fig1} shows the overall procedure of FO-MoE. Our main contributions can be summarized as follows:
 
\begin{itemize}
    \item We introduce \textit{FO-MoE}, the first fairness-oriented MoE designed explicitly for medical VLM application, which enhances task relevance while mitigating bias in extracted features.
    \item To consider both fairness and efficiency, \textit{FOL}, a novel fairness-aware loss, is proposed to integrate both distance and dispersion metrics, ensuring robust fairness. 
    \item We present Fair-MoE, a new framework consisting of FO-MoE and FOL for advancing fairness in medical VLM, bridging the gap in existing research and setting a foundation for future developments in this domain.
   \item To validate the fairness and effectiveness of our methods, we validate it on the Harvard-FairVLMed database with comprehensive experiments, including multiple ablation studies, demonstrating significant improvements in both accuracy and fairness metrics. 
\end{itemize}

\section{Relate Work}

\subsection{Fairness in Machine Learning}
Ensuring fair decision-making by machine learning algorithms means migrating disparities in outcomes between different demographic groups, ensuring equitable results regardless of attributes such as race, gender, or ethnicity\cite{parraga2023fairness,luo2024fairclip,mehrabi2021survey}. Vision models, which process visual data, have been extensively studied for fairness issues in both natural and medical image domain. For instance, methods like Dr-Fairness\cite{rohdr} dynamically balance real and generated data for equitable outcomes, while Debiased Self-Attention (DSA)\cite{qiang2023fairness} identifies and masks spurious features in Vision Transformers using adversarial examples. In medical image analysis, studies have shown that models trained on chest X-ray datasets often encode protected attributes even when not explicitly labeled, leading to disparities in detection accuracy across groups\cite{glocker2023algorithmic}. Similarly, foundation models exhibit performance gaps across demographic subgroups due to imbalanced pre-training datasets\cite{khan2023fair}. In addition, the fairness of traditional machine learning, where features are predefined and models often perform classification tasks, has been widely studied. These works introduce fairness metrics and bias mitigation strategies to address inequities in unimodal data, such as images or text\cite{germino2024fairmoe,sharma2022feamoe}. However, these methods are typically limited to single-modality data and fail to address the unique challenges posed by multimodal systems like Vision-Language Models (VLMs), where visual and textual data interact and can compound fairness issues.

\subsection{Vision Language Model}
Vision-Language Models (VLMs) are designed to process and integrate visual and textual data together, enabling them to tackle complex multimodal tasks. One of the most popular and pioneering models of these is CLIP, which utilizes contrastive learning to align image and text embeddings, allowing it to generalize across diverse tasks without task-specific fine-tuning. In recent years, because of its ability to process both medical images and associated textural data for comprehensive analysis and reducing the need of time-intensive manual annotation, CLIP has gained significant attention in medical applications. Many studies have explored its potential in the medical imaging domain and proposed methods to enhance its efficiency and explainability. For example, parameter-efficient transfer learning (PETL)\cite{liu2023parameter} framework utilizes lightweight adapters integrated into the pre-trained CLIP architecture to fine-tune efficiently without modifying the entire model. Another notable approach leverages ChatGPT alongside pre-trained CLIP to generate more interpretable and comprehensive diagnostic reports based on image predictions\cite{liu2023chatgpt}. While optimizing the performance and explainability of VLMs is essential, addressing fairness issues is equally critical. To tackle fairness challenges in VLMs, the first fairness-oriented medical VLM dataset, Harvard-FairVLMed\cite{luo2024fairclip}, was introduced in 2024. Built upon this dataset, FairCLIP\cite{luo2024fairclip} was proposed to mitigate biases in protected attribute distributions using Sinkhorn distance, extending CLIP’s capabilities to address fairness. Although FairCLIP represents a step forward in fairness research for VLMs, it relies solely on Sinkhorn distance to mitigate biases. However, its retention of the original CLIP architecture restricts the model’s capacity to effectively learn and address underlying unfair information.

\section{Method}
\textcolor{red}{Fig. \ref{fig1}} demonstrates workflow of proposed Fair-MoE. Several attention blocks are stacked together to extract features from text and image. Multi-head attention computed in the last attention block is fed into FO-MoE consisting of patch embedding-based FO-MoE and feature-based FO-MoE to get fair text and image features. Finally, similarity between fair text and image features and proposed novel loss function FOL are utilized to optimize the model.  % Text encoder and image encoder extract features from texts and images. Several %Both text encoder and image encoder contains FO-MoE discussed in Section \ref{FO-MoE} consisted of Patch embedding-based .    

\subsection{Fairness-Oriented Mixture of Expert (FO-MoE)}
\label{FO-MoE}
Recent fairness-oriented VLMs in the medical domain, based on CLIP, aim to achieve fairness by minimizing distances between different groups' distributions \cite{luo2024fairclip}. However, CLIP's architecture may inadvertently learn biased information, as it processes all inputs indiscriminately through its encoders, limiting its capacity to ensure unbiased learning.

%To enhance learning ability and avoid learning biased content directly from images, the architecture has been modified by implementing FO-MoE in both image and text encoder. This modification replaces the MLP layer in the last attention block for each encoder with a patch embedding-based MoE layer and places a feature-based MoE layer after the encoders. For clarity, the description below focuses solely on the implementation of FO-MoE in image encoder. However, the same approach is applied in text encoder, with the only difference being that the input is textual.

To enhance learning ability and avoid learning biased content directly from images, the architecture has been modified by implementing FO-MoE in both image and text encoder. This modification replaces the MLP layer in the last attention block for each encoder with a patch \textit{embedding-based MoE} layer and places a \textit{feature-based MoE} layer after the encoders. 
Each MoE layer comprises multiple experts, which are MLPs designed to capture and learn distinct aspects of information from the inputs. Additionally, the input passes through a gating mechanism, also implemented as a MLP, which assigns a weight to each expert. The weight indicates the likelihood that an expert should process the input, and output is aggregated by weighted summing outputs of experts.

\begin{figure*}[htbp]
    \centerline{\includegraphics[width=\linewidth]{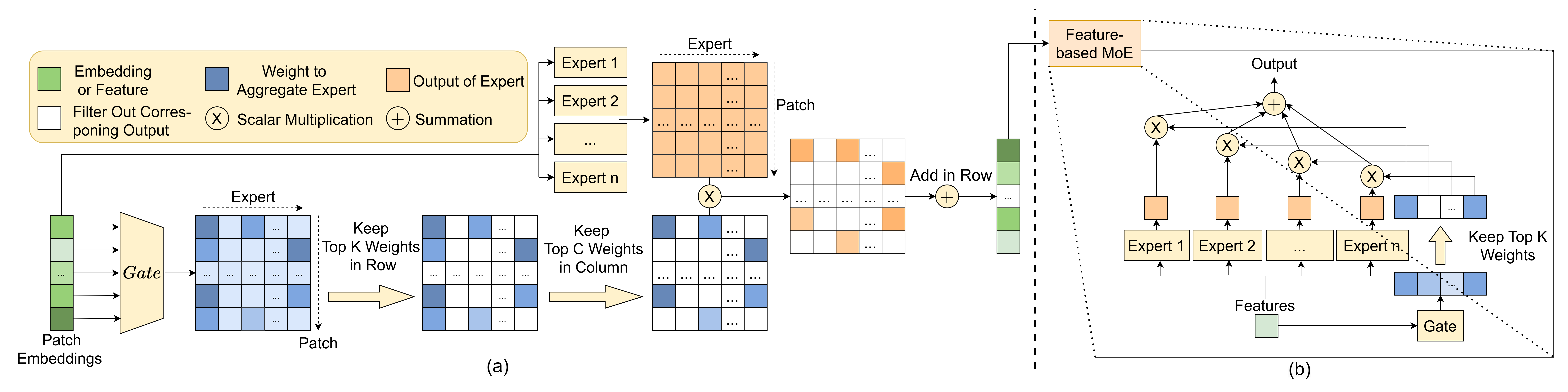}}
    \caption{(a) is an illustration of embedding-based MoE. Blue $matrix_{i, j}$ represents the weight to aggregate output of $j$-th expert for $i$-th embedding. Orange $matrix_{i, j}$ represents the output of $j$-th expert when inputs $i$-th embedding. (b) is an illustration of feature-based MoE's structure. For both images,
    green represents patch embedding or feature. Blue represents weight output by gate to aggregate outputs of experts. Orange represents output of expert. White denotes zero, indicating that corresponding output of expert has been filtered out. A darker color indicates a higher number.} %signifying stronger influence or relevance.}
    \label{fig3}
\end{figure*}

\textit{Embedding-based MoE:} Let $I^1$ denote the input to the embedding-based MoE, which is the output processed through preceding blocks before the final one of the encoder. And $I^1 \in R^{(N+1) \times D} $, where $N$ is the number of patches in input, and $D$ is the dimension of patch embeddings. Then, in the embedding-based MoE, the weight matrix, which includes the weight assigned by gating mechanism to each expert for each embedding, is defined as $ W^1 = softmax((G^1(I^1)))$, $\hat{W^1} = Top_c(Top_r(W^1, k^1), \alpha)$, where $G^1$ is the gating function that assigns each expert a weight for all embeddings. Formally, it can be written as: $G^1: R^{(N+1) \times D} \xrightarrow{}  R^{(N+1) \times M^1}$, where $M^1$ denote the number of experts. The $softmax$ function transforms the output into a probability space, showing likelihood of each expert being suitable to process the input. The $Top_r$ function is utilized to boost performance by adopting a sparse MoE approach, which preserves the $k^1$ largest weights in each row of the input while setting the other weights to zero \cite{riquelme2021scaling}. To filter out biased path embeddings, a capacity $C$ is introduced, representing the number of embeddings that an expert can process. $Top_c$ here is used to achieve it by only keeping highest $\alpha = \frac{C(N+1)k^1}{M^1}$ weights in column. Meanwhile, embeddings also need to pass through experts. Let $I^2\in R^{(N+1) \times \hat{D}}$ be the aggregates output of experts. Formally, row $a$ of $I^2$ can be written as: $I^2_a = \sum_{b = 0}^{M^1-1}\hat{W}_{a, b}^1E^1_{b}(I^1_a)$. $E^1_{b}(x)$ denotes an expert in the patch embedding-based MoE, which is a two layers MLP with an activation function $E^1_{b}(x) = \widetilde{T^1_{b}}\sigma(\widetilde{W^1_{b}}x), b \in [M^1] $. \textcolor{red}{Fig \ref{fig3} (a)} illustrates the workflow of embedding-based MoE.

\textit{Feature-based MoE:} Following vision transformer\cite{dosovitskiy2020vit} , the first patch embeddings $I^2_0\in R^D$ are selected as a feature vector. These feature vectors will be sent to a feature-based MoE with $M^2$ experts that further eliminates biased information to get the fair feature. The structure of it is shown in \textcolor{red}{Fig \ref{fig3} (b)}. The output $W^2=Top_r(softmax(G^2(I^2_0)), k^2)$ that keeps highest $k^2$ weights from gates $G^2: R^D \xrightarrow{} R^{M^2}$ is used to aggregate outputs from experts to obtain a more fair visual feature vector $I^3 = \sum_{b = 0}^{M^2-1}\hat{W}_{{ b}}^2E^2_{b}(I^2_0)$. $\hat{W}_{b}^2$ is a scalar indicate of $b$ th element in $\hat{W}_{b}^2$. $E^2_{b}(x)$ denotes $b$th experts in feature-based MoE.

\subsection{Fairness-Oriented Loss (FOL)}
Optimizing the variance of weights to aggregate outputs from different experts enhances the learning capacity of the Mixture-of-Experts (MoE) model by achieving load balance across the experts \cite{lou2021cross}. Furthermore, variance, as a measure of distribution dispersion, plays a critical role in fairness. By optimizing the variance differences between distributions of protected attribute groups, disparities in these distributions can be reduced. Building on this principle, we can improve existing fairness loss functions which focus on optimizing distance between distribution of protected attribute groups to decrease disparities among different distributions of protected attribute groups\cite{luo2024fairclip,tian2024fairvit} by  leveraging variance utilized to load balance loss to develop a new fair loss, FOL, that takes both distance and dispersion into account.  

In FOL, the weight output by a gate for a certain expert is selected as a random variable. To estimate the variance, we sample $N$ data pairs from whole dataset and $N$ data pairs from protected attribute group. Take embedding-based MoE of image as an instance, inputting single image data $I^1$ gives weight $\hat{W}^1$, where $\hat{W}^1_{a,b}$ represents weight of integrating output from expert $b$ when input $a$th patch embedding. To estimate the variance, all weights $\hat{W}^1$ computed from data sampled from the whole dataset are stacked together and denoted as $O_N$. Meanwhile, all weights $\hat{W}^1$ computed from data sampled from certain protected attribute group $p$ are stacked together and denoted as $O_{N|p}$. Variance difference between $i$th column of $O_N$ and $O_{N|p}$ demonstrates dispersion between different attribute's distribution in weight of expert $i$. To optimize dispersion between different attribute's distribution, model should optimize difference of variance for all experts. Thus, loss for embedding-based MoE of image is $F_{EI} = \sum_{p \in P}\sum_{j=0}^{M^1-1} (Var(O_{N_j})-Var(O_{N|p_j}))^2$ where $O_{N_j}$, $O_{N|p_j}$ denote $j$th column of $O_{N}$, $O_{N|p}$ which denote all expert $j$'s weights. $Var(\cdot)_j$ means compute variance of input. $P$ is a set of groups for certain attribute. In the same way, loss for embedding-based MoE of text $F_{ET}$, loss for feature-based MoE of image $F_{FI}$ and loss for feature-based MoE of text $F_{FT}$ can be gotten. Finally, FOL is defines as $FOL=F_{EI}+F_{ET}+F_{FI}+F_{FT}+L_{distance}$, where $L_{distance}$ is Sinkhorn distance loss\cite{peyre2019computational}.

\section{Experiment}
\subsection{Experimental Setup}
Experiments are conducted on the Harvard-FairVLMed database \cite{luo2024fairclip}, which comprises 7,000 training samples, 1,000 validation samples, and 2,000 test samples. Each sample in the database includes an SLO fundus image, accompanying clinical notes, labels of image-text pairs and protected attributes such as the patient's race, gender (GEN), ethnicity (ETH), and language (LAN). To demonstrate the fairness capabilities of the Fair-MoE model compared with the baselines, the training protocol was aligned with that of the FairCLIP \cite{luo2024fairclip}. All experiments were conducted on an NVIDIA GeForce RTX 3090 GPU. To comprehensively evaluate the fairness and performance of FairMoE, four metrics are employed. The Area Under the Curve (AUC) is utilized to measure the model's overall performance. To assess fairness, the Demographic Parity Difference (DPD) and Equal Opportunity Difference (EOD) are used, providing insights into potential biases within the model. Additionally, the Equity-Scaled AUC (ES-AUC) is introduced to evaluate the trade-off between performance and fairness. This metric is particularly critical, as improving fairness may result in a reduction in model performance. The ES-AUC quantifies whether enhancements in fairness achieve an acceptable balance with performance. The detailed definitions and explanations of these four metrics are provided below:%utilizing the following evaluation metrics:
\begin{itemize}
    \item (1) Area Under the Receiver Operating Characteristic Curve (AUC) quantifies a model's effectiveness in ranking positive samples higher than negative ones, serving as a prevalent performance metric in medical diagnostics. 
    \item (2) Demographic Parity Difference (DPD) measures the fairness of the model by comparing the probability of a good outcome across different demographic groups. For all group $a$ and $b$ of a protected attribute $s$, it is defined as $DPD_s = |\max_{a}P(\hat{y}=1| G=a, y=1) - \min_{b}P(\hat{y}=1 | G=b, y=1)|$, where $a \neq b$. 
    \item (3) Difference in Equalized Odds (EOD) is another fairness metric that considers both the true positive rates (TPR) and false positive rates (FPR) across groups. EOD of an attribute $s$ is calculated as $EOD_s = \max_{a,b \in s, a \neq b}(|P(\hat{y}=1 | G=a, y=1) - P(\hat{y}=1 | G=b, y=1)|, |P(\hat{y}=1 | G=a, y=0) - P(\hat{y}=1 | G=b, y=0)|)$.
    \item (4) Equity-Scaled AUC (ES-AUC) assesses how equally the AUC of a model is distributed across different demographic groups. Formally, ES-AUC of an attribute $s$ is $ES-AUC_s = \frac{AUC_s}{1 + \sum_a{|AUC_s - AUC_{s,a}|}}$ where $AUC_s$ is the overall AUC of attribute $s$ and $AUC_{s,a}$ is the AUC of a group $a \in s$.
\end{itemize}

\subsection{Comparison with Baselines}
\begin{table}[t]
\caption{Main Results of Experiments. The green text highlights our method.}
\centering
\resizebox{0.5\textwidth}{!}{
\begin{tabular}{c|c|cccc} 
\hline
Attr.                  & Model        & ES-AUC               & AUC                  & DPD                 & EOD                   \\ 
\hline
\multirow{6}{*}{Race}      & CLIP/b16     & 62.67$\pm$3.15          & 67.70$\pm$3.13          & 14.57$\pm$3.77         & 18.47$\pm$5.12           \\
                           & CLIP/l14     & 66.83$\pm$2.19          & 70.63$\pm$2.98          & 11.69$\pm$3.85         & 15.13$\pm$2.66           \\
                           & FairCLIP/b16 & 61.17$\pm$1.87          & 67.47$\pm$1.16          & 10.16$\pm$10.05        & 11.44$\pm$11.07          \\
                           & FairCLIP/l14 & 67.53$\pm$4.26          & 71.57$\pm$2.94          & 16.01$\pm$5.87         & 17.03$\pm$3.74           \\
                           & \textcolor{darkgreen}{FairMoE/b16}  & 69.63$\pm$1.21          & 71.93$\pm$0.90          & 7.25$\pm$5.13          & 7.43$\pm$3.04            \\
                           & \textcolor{darkgreen}{FairMoE/l14}  & \textbf{72.53}$\pm$1.07 & \textbf{73.93}$\pm$0.97 & \textbf{2.63}$\pm$0.65 & \textbf{4.25}$\pm$0.75   \\ 
\hline\hline
\multirow{6}{*}{GEN}    & CLIP/b16     & 63.30$\pm$2.73          & 67.70$\pm$3.13          & 2.79$\pm$1.49          & 7.52$\pm$4.78            \\
                           & CLIP/l14     & 66.30$\pm$2.63          & 70.63$\pm$2.98          & 3.13$\pm$2.60          & 7.56$\pm$3.54            \\
                           &  FairCLIP/b16 & 64.43$\pm$1.86          & 68.47$\pm$2.26          & 2.50$\pm$1.47           & 4.98$\pm$3.74            \\
                           & FairCLIP/l14 & 67.37$\pm$1.62          & 70.80$\pm$1.84           & 2.11$\pm$1.81          & 5.24$\pm$1.46            \\
                           & \textcolor{darkgreen}{FairMoE/b16}  & 68.07$\pm$0.96          & 71.97$\pm$1.16          & \textbf{1.91}$\pm$1.02 & \textbf{3.53}$\pm$0.90   \\
                           & \textcolor{darkgreen}{FairMoE/l14}  & \textbf{69.97}$\pm$3.39 & \textbf{74.97}$\pm$2.90 & 2.94$\pm$1.60          & 7.33$\pm$2.55            \\ 
\hline\hline
\multirow{6}{*}{ETH} & CLIP/b16     & 64.87$\pm$2.26          & 70.63$\pm$0.90          & 7.53$\pm$2.96          & 14.83$\pm$3.01           \\
                           & CLIP/l14    & 64.13$\pm$1.58          & 69.37$\pm$1.04          & 8.74$\pm$0.41          & 9.13$\pm$0.69            \\
                           &  FairCLIP/b16 & 61.43$\pm$1.05          & 67.33$\pm$1.33          & 10.54$\pm$1.52         & 17.93$\pm$4.01           \\
                           & FairCLIP/l14 & 64.23$\pm$1.11          & 69.23$\pm$0.92          & 15.37$\pm$2.17         & 15.77$\pm$3.17           \\
                           & \textcolor{darkgreen}{FairMoE/b16}  & 65.17$\pm$2.44          & 69.77$\pm$0.49          & \textbf{8.52}$\pm$3.19 & \textbf{8.42}$\pm$2.77   \\
                           & \textcolor{darkgreen}{FairMoE/l14}  & \textbf{67.10}$\pm$4.70  & \textbf{72.80}$\pm$2.54 & 8.79$\pm$2.91          & 13.90$\pm$5.86           \\ 
\hline\hline
\multirow{6}{*}{LAN}  & CLIP/b16     & 60.10$\pm$3.84          & 67.70$\pm$3.13          & 13.50$\pm$3.96         & 16.40$\pm$9.56           \\
                           & CLIP/l14     & 59.90$\pm$2.01          & 69.37$\pm$1.04          & 17.27$\pm$0.74         & 20.17$\pm$6.09           \\
                           &  FairCLIP/b16 & 57.97$\pm$0.65          & 68.07$\pm$0.57          & 10.96$\pm$4.04         & 14.25$\pm$9.09           \\
                           & FairCLIP/l14 & 63.57$\pm$1.97          & 72.40$\pm$1.84          & 8.21$\pm$1.99          & \textbf{11.00}$\pm$1.25  \\
                           & \textcolor{darkgreen}{FairMoE/b16}  & 63.60$\pm$1.85          & \textbf{73.87}$\pm$1.62 & \textbf{7.48}$\pm$4.56 & 12.30$\pm$2.65           \\
                           & \textcolor{darkgreen}{FairMoE/l14}  & \textbf{63.80}$\pm$1.28 & 71.37$\pm$2.10          & 15.67$\pm$2.99         & 23.63$\pm$14.40          \\
\hline
\end{tabular}
}
\label{tabel1}
\end{table}
To evaluate the performance and fairness of Fair-MoE in medical images, two SoTA fairness-aware VLMs, i.e., Vanilla and FairCLIP, are chosen as the baselines. \textcolor{red}{Table \ref{tabel1}} demonstrates the results of comparing Fair-MoE with CLIP and the SOTA fair medical vision language model Fair-CLIP.For ES-AUC, which takes both effectiveness and fairness into account, Fair-MoE outperforms all baselines in all protected attributes. For attribute race, Fair-MoE outperforms baselines $5.00\%$ in ES AUC. For AUC that measures effectiveness of model, Fair-MoE also outperforms all baselines in all protected attributes. For attribute gender, Fair-MoE achieves $4.91\%$ improvement in AUC. For DPD and EOD that measure the fairness of model, results of DPD show that Fair-MoE achieves better fairness than baselines in all attributes. Besides, results of EOD show that Fair-MoE achieves better fairness than baselines in attributes of race, gender, and ethnicity. The results prove that in addition to achieving a better trade-off between effectiveness and fairness, Fair-MoE can both improve effectiveness and fairness. The parameter counts for CLIP/B16, FairCLIP/B16, and FairMoE/B16 are approximately 200M, while those for CLIP/L14, FairCLIP/L14, and FairMoE/L14 are approximately 500M. These results demonstrate that FairMoE achieves improvements in both accuracy and fairness without a significant increase in model parameter count, maintaining computational efficiency while enhancing performance and fairness.

\begin{table}[t]
\caption{Results of ablation study of Fairness-Oriented Mixture of Expert, FOM denotes Fairness-Oriented Mixture of Expert.}
\centering
\resizebox{\linewidth}{!}{
\begin{tabular}{c|c|cccc} 
\hline
Attr.                 & Model              & ES-AUC               & AUC                  & DPD                 & EOD                  \\ 
\hline
\multirow{4}{*}{Race}      & FairCLIP/b16       & 61.17$\pm$1.87          & 67.47$\pm$1.16          & 10.16$\pm$10.05        & 11.44$\pm$11.07         \\
                           & FairCLIP/l14       & 67.53$\pm$4.26          & 71.57$\pm$2.94          & 16.01$\pm$5.87         & 17.03$\pm$3.74          \\
                           & FairCLIP/b16 w. FOM & \textbf{69.97}$\pm$2.60 & \textbf{72.67}$\pm$1.16 & \textbf{3.19}$\pm$2.04 & \textbf{9.48}$\pm$3.74  \\
                           & FairCLIP/l14 w. FOM & 65.53$\pm$4.74          & 67.10$\pm$3.85          & 13.37$\pm$8.32         & 13.24$\pm$6.93          \\ 
\hline\hline
\multirow{4}{*}{GEN}    & FairCLIP/b16       & 64.43$\pm$1.86          & 68.47$\pm$2.26          & 2.50$\pm$1.47          & \textbf{4.98}$\pm$3.74  \\
                           & FairCLIP/l14        & 67.37$\pm$1.62          & 70.80$\pm$1.84          & \textbf{2.11}$\pm$1.81 & 5.24$\pm$1.46           \\
                           & FairCLIP/b16 w. FOM & \textbf{67.63}$\pm$1.82 & \textbf{71.37}$\pm$2.30 & 3.14$\pm$1.14          & 7.70$\pm$1.66           \\
                           & FairCLIP/l14 w. FOM & 64.33$\pm$1.83          & 68.67$\pm$1.87          & 4.25$\pm$3.08          & 7.22$\pm$3.17           \\ 
\hline\hline
\multirow{4}{*}{ETH} & FairCLIP/b16       & 61.43$\pm$1.05          & 67.33$\pm$1.33          & 10.54$\pm$1.52         & 17.93$\pm$4.01          \\
                           & FairCLIP/l14        & 64.23$\pm$1.11          & 69.23$\pm$0.92          & 15.37$\pm$2.17         & 15.77$\pm$3.17          \\
                           & FairCLIP/b16 w. FOM & \textbf{66.70}$\pm$3.30 & 69.17$\pm$2.57          & \textbf{6.94}$\pm$4.40 & \textbf{9.58}$\pm$4.11  \\
                           & FairCLIP/l14 w. FOM & 66.57$\pm$3.94          & \textbf{71.60}$\pm$2.71 & 11.72$\pm$2.35         & 15.97$\pm$0.78          \\ 
\hline\hline
\multirow{4}{*}{LAN}  & FairCLIP/b16       & 57.97$\pm$0.65          & 68.07$\pm$0.57          & 10.96$\pm$4.04         & 14.25$\pm$9.09          \\
                           & FairCLIP/l14        & \textbf{63.57}$\pm$1.97 & 72.40$\pm$1.84          & \textbf{8.21}$\pm$1.99 & 11.00$\pm$1.25          \\
                           & FairCLIP/b16 w. FOM & 62.47$\pm$2.53          & \textbf{72.53}$\pm$1.09 & 11.97$\pm$3.87         & 23.20$\pm$2.12          \\
                           & FairCLIP/l14 w. FOM & 62.33$\pm$0.68          & 65.17$\pm$0.90          & 10.43$\pm$0.42         & \textbf{9.65}$\pm$3.05  \\
\hline
\end{tabular}
}
\label{tabel2}
\end{table}

\begin{table}
\caption{Results of Ablation study of Fairness-Oriented Loss (FOL). The green texts highlights our method}
\centering
\resizebox{\linewidth}{!}{
\begin{tabular}{c|c|cccc} 
\hline
Attr.               & Model                & ES-AUC               & AUC                  & DPD                 & EOD                  \\ 
\hline
\multirow{4}{*}{Race}      & FairMoE/b16 w/o FOL & 69.97$\pm$2.60          & 72.67$\pm$1.16          & 3.19$\pm$2.04          & 9.48$\pm$3.74           \\
                           & FairMoE/l14 w/o FOL & 65.53$\pm$4.74          & 67.10$\pm$3.85          & 13.37$\pm$8.32         & 13.24$\pm$6.93          \\
                           & \textcolor{darkgreen}{FairMoE/b16}          & 69.63$\pm$1.21          & 71.93$\pm$0.90          & 7.25$\pm$5.13          & 7.43$\pm$3.04           \\
                           & \textcolor{darkgreen}{FairMoE/l14}          & \textbf{72.53}$\pm$1.07 & \textbf{73.93}$\pm$0.97 & \textbf{2.63}$\pm$0.65 & \textbf{4.25}$\pm$0.75  \\ 
\hline\hline
\multirow{4}{*}{GEN}    & FairMoE/b16 w/o FOL & 67.63$\pm$1.82          & 71.37$\pm$2.30          & 3.14$\pm$1.14          & 7.70$\pm$1.66           \\
                           & FairMoE/l14 w/o FOL & 64.33$\pm$1.83          & 68.67$\pm$1.87          & 4.25$\pm$3.08          & 7.22$\pm$3.17           \\
                           & \textcolor{darkgreen}{FairMoE/b16}          & 68.07$\pm$0.96          & 71.97$\pm$1.16          & \textbf{1.91}$\pm$1.02 & \textbf{3.53}$\pm$0.90  \\
                           & \textcolor{darkgreen}{FairMoE/l14}          & \textbf{69.97}$\pm$3.39 & \textbf{74.97}$\pm$2.90 & 2.94$\pm$1.60          & 7.33$\pm$2.55           \\ 
\hline\hline
\multirow{4}{*}{ETH} & FairMoE/b16 w/o FOL & 66.70$\pm$3.30          & 69.17$\pm$2.57          & \textbf{6.94}$\pm$4.40 & 9.58$\pm$4.11           \\
                           & FairMoE/l14 w/o FOL & 66.57$\pm$3.94          & 71.60$\pm$2.71          & 11.72$\pm$2.35         & 15.97$\pm$0.78          \\
                           & \textcolor{darkgreen}{FairMoE/b16}          & 65.17$\pm$2.44          & 69.77$\pm$0.49          & 8.52$\pm$3.19          & \textbf{8.42}$\pm$2.77  \\
                           & \textcolor{darkgreen}{FairMoE/l14}          & \textbf{67.10}$\pm$4.70 & \textbf{72.80}$\pm$2.54 & 8.79$\pm$2.91          & 13.9$\pm$5.86           \\ 
\hline\hline
\multirow{4}{*}{LAN}  & Fair-MoE/b16 w/o FOL & 62.47$\pm$2.53          & 72.53$\pm$1.09          & 11.97$\pm$3.87         & 23.20$\pm$2.12          \\
                           & Fair-MoE/l14 w/o FOL & 62.33$\pm$0.68          & 65.17$\pm$0.90          & 10.43$\pm$0.42         & \textbf{9.65}$\pm$3.05  \\
                           & \textcolor{darkgreen}{FairMoE/b16}          & 63.60$\pm$1.85          & \textbf{73.87}$\pm$1.62 & \textbf{7.48}$\pm$4.56 & 12.30$\pm$2.65          \\
                           & \textcolor{darkgreen}{FairMoE/l14}          & \textbf{63.80}$\pm$1.28 & 71.37$\pm$2.10          & 15.67$\pm$2.99         & 23.63$\pm$14.4          \\
\hline
\end{tabular}
}
\label{tabel3}
\end{table}

\begin{table}[t]
\caption{Results of Ablation study on $F_{EI}$, $F_{ET}$, $F_{FI}$ and $F_{FT}$ in ES-AUC.}
\resizebox{\linewidth}{!}{
\begin{tabular}{c|cccc}
\hline
Model               & Race & Gender & Ethnicity & Language \\ \hline
\textcolor{darkgreen}{FairMoE/b16}         & \textbf{70.9} & \textbf{70.4}   & \textbf{70.7}      & \textbf{66.1}     \\
w/o loss of $F_{EI}$/b16 & 62.2 & 65.4   & 58.7      & 60       \\
w/o loss of $F_{ET}$/b16 & 62.4 & 62     & 64.4      & 58.3     \\
w/o loss of $F_{FI}$/b16 & 70.4 & 56.5   & 61.9      & 61.7     \\
w/o loss of $F_{FT}$/b16 & 60.9 & 69.8   & 62.2      & 48.7     \\ \hline
\textcolor{darkgreen}{FairMoE/l14}         & \textbf{74.0}   & \textbf{69.5}   & \textbf{73.4}      & \textbf{64.1}     \\
w/o loss of $F_{EI}$/l14 & 71.4 & 62.8   & 64.7      & 62.5     \\
w/o loss of $F_{ET}$/l14 & 64.3 & 59.2   & 69.6      & 62.4     \\
w/o loss of $F_{FI}$/14  & 69.2 & 63.0     & 63.4      & 59.6     \\
w/o loss of $F_{FT}$/l14 & 69.3 & 64.6   & 70.1      & 59.7     \\ \hline
\end{tabular}
}
\label{tabel4}
\end{table}

\begin{table}
\caption{Results of Ablation study on Embedding-based MoE (EM) and Feature-based MoE (FM) in ES-AUC.}
\resizebox{\linewidth}{!}{
\begin{tabular}{c|cccc}
\hline
Model              & Race & Gender & Ethnicity & Language \\ \hline
\textcolor{darkgreen}{FairMoE/b16}        & \textbf{70.9} & \textbf{70.4}   & \textbf{70.7}      & \textbf{66.1}     \\
FairMoE/b16 w/o EM & 66.2 & 68.1   & 53.5      & 62.9     \\
FairMoE/b16 w/o FM & 64.0 & 66.5   & 66.3      & 61.0     \\ \hline
\textcolor{darkgreen}{FairMoE/l14}        & \textbf{74.0} & \textbf{69.5}   & \textbf{73.4}      & \textbf{64.1}     \\
FairMoE/l14 w/o EM & 68.6 & 66.9   & 62.0      & 62.2     \\
FairMoE/l14 w/o FM & 72.2 & 65.4   & 72.1      & 60.8     \\ \hline
\end{tabular}
}
\label{tabel5}
\end{table}

\begin{table}[t]
\caption{Results of Ablation study on MoE modules in Text and Image in ES-AUC.}
\resizebox{\linewidth}{!}{
\begin{tabular}{c|cccc}
\hline
Model             & Race & Gender & Ethnicity & Language \\ \hline
\textcolor{darkgreen}{FairMoE/b16}       & \textbf{70.9} & \textbf{70.4}   & \textbf{70.7}      & \textbf{66.1}     \\
w/o Text MoE/b16  & 66.8 & 67.2   & 61.3      & 63.6     \\
w/o Image MoE/b16 & 69.4 & 66.8   & 64.6      & 54.8     \\ \hline
\textcolor{darkgreen}{FairMoE/l14}       & \textbf{74.0} & \textbf{69.5}   & \textbf{73.4}      & \textbf{64.1}     \\
w/o Text MoE/l14  & 72.1 & 61.3   & 64.0      & 63.8     \\
w/o Image MoE/l14 & 66.8 & 65.3   & 64.3      & 58.7     \\ \hline
\end{tabular}
}
\label{tabel6}
\end{table}

\subsection{Ablation Study}
To analyze performance of FO-MoE and FOL, two ablation studies are implemented. Firstly, to assess performance of FO-MoE, architecture of Fair-CLIP is changed to architecture of FO-MoE. \textcolor{red}{Table \ref{tabel2}} demonstrates results of ablation study of FO-MoE. Utilizing FO-MoE achieves higher AUC for all attributes demonstrates and gains $1.1\%$ improvements in race, demonstrating that MoE can enhance FO-MoE's learning capabilities and enable FO-MoE to achieve advanced effectiveness. For majority of attributes, applying FO-MoE can achieve higher ES-AUC, which indicates that adding FO-MoE achieves a better trade-off between effectiveness and fairness than FairCLIP. For attribute race and ethnicity, applying FO-MoE can both improve effectiveness and fairness. This phenomenon proves FO-MoE's ability to filter out
bias patch embedding and extract more fair task-relevant information. An intriguing phenomenon is observed in the ViT/L14 architecture, where incorporating FO-MoE into FairCLIP occasionally results in a performance drop. This behavior can be attributed to the relatively small size of the Harvard-FairVLMed dataset (~8k samples), which increases the risk of overfitting in the ViT/L14 architecture. Furthermore, the absence of a tailored loss function for MoE models, such as the proposed FOL, exacerbates this overfitting issue when FO-MoE is introduced. This finding highlights the critical importance of designing specialized loss functions, like FOL, to address the overfitting challenges posed by MoE-based architectures.

Secondly, to assess performance of FOL, we remove FOL from Fair-MoE, \textcolor{red}{Table \ref{tabel3}} shows how removing FOL from Fair-MoE will affect performance of Fair-MoE. In the case of removing FOL for all four attributes, metrics that measure effectiveness and fairness deteriorate significantly. Removing FOL leads to a drop of $2.56\%$ in AUC for race and $2.34\%$ in $ES-AUC$ for gender. The drop in performance proves that just minimizing the distance between different attributes' distribution is not enough. Thus, optimizing difference between dispersion of attributes' distribution is indispensable to achieve a leap in both effectiveness and fairness. In addition, optimizing dispersion can improve stability of MoE, letting Fair-MoE better filter out bias patch embedding and utilize its supreme learning capacities to extract fair feature. The results of two ablation studies prove the effectiveness of FO-MoE and FOL.

Thirdly, the proposed FOL consists of five components: $F_{EI}$, $F_{ET}$, $F_{FI}$, $F_{FT}$, and $L_{distance}$. While $L_{distance}$ is widely adopted for improving fairness, the other four components are introduced for the first time in this work. To evaluate the effectiveness of these components, we perform an ablation study by individually removing $F_{EI}$, $F_{ET}$, $F_{FI}$, and $F_{FT}$ from Fair-MoE. The results of this study, presented in \textcolor{red}{Table \ref{tabel4}}, highlight the necessity of utilizing all four components. The findings demonstrate that removing any of $F_{EI}$, $F_{ET}$, $F_{FI}$, or $F_{FT}$ results in a decline in ES-AUC, a metric quantifying the trade-off between fairness and performance. These components are specifically designed to guide the corresponding embedding-based FO-MoEs and feature-based FO-MoEs in filtering out bias and irrelevant features while ensuring better load balancing. Eliminating any of these components compromises the fairness and performance of the corresponding MoE layers, as evidenced by the performance drops shown in \textcolor{red}{Table \ref{tabel4}}.

Fourthly, FO-MoE is composed of two key components: the embedding-based MoE, integrated into the last attention block, and the feature-based MoE, positioned after the encoder. To demonstrate the effectiveness of these components, we conduct an ablation study by separately removing the embedding-based MoE and feature-based MoE from Fair-MoE. The results, presented in \textcolor{red}{Table \ref{tabel5}}, validate the contributions of both components. Removing the embedding-based MoE results in a noticeable decline in the trade-off between performance and fairness across all four attributes. This is because, without the embedding-based MoE, Fair-MoE loses its capability to filter out bias and task-irrelevant patch embeddings. Consequently, the model is more prone to incorrectly leveraging these biased and irrelevant embeddings during decision-making. Similarly, removing the feature-based MoE also reduces the trade-off between performance and fairness. This highlights the feature-based MoE's critical role in enhancing the model's learning capacity, enabling it to focus on task-relevant information while minimizing the extraction of bias and task-irrelevant features. These findings underscore the importance of both embedding-based and feature-based MoEs in achieving a balance between fairness and performance, as evidenced by the results in \textcolor{red}{Table \ref{tabel5}}.

%Fifthly, to explore whether Fair-MoE is conducive to both image and text modality, we remove FO-MoE and FOL in text modality to test effectiveness of Fair-MoE in text modality and we remove FO-MoE and FOL in image modality to test effectiveness of Fair-MoE in image modality. \textcolor{red}{Table \ref{tabel6}} demonstrates results of ablation study. When either removing MoE in text modality and image modality, trade-off between performance and fairness drops in all four attributes. Results prove that when removing Fair-MoE in text or image modality, the feature extracts from corresponding text or image contains more bias but task-irrelevant information. These findings show that Fair-MoE's effectiveness in both text modality and image modality. 
Fifthly, to investigate whether Fair-MoE benefits both image and text modalities, we conduct an ablation study by removing FO-MoE and FOL from the text modality to evaluate the effectiveness of Fair-MoE in text processing, and similarly, by removing FO-MoE and FOL from the image modality to evaluate its effectiveness in image processing. \textcolor{red}{Table \ref{tabel6}} presents the results of this study. The results show that removing MoE from either the text modality or the image modality leads to a drop in the trade-off between performance and fairness across all four attributes. This indicates that without Fair-MoE, the features extracted from the corresponding text or image contain more bias and task-irrelevant information. These findings confirm the effectiveness of Fair-MoE in enhancing trade-off between fairness and performance in both text and image modalities.

\section{conclusion}
We propose a new algorithm Fair-MoE that can both improve effectiveness and fairness in medical VLMs. 
% Fair-MoE applies FO-MoE to learn unbias feature and filter out bias information. 
Fair-MoE includes two key components: \textit{FO-MoE and FOL}. \textit{FO-MoE} is designed to learn unbiased features and filter out biased information.
Meanwhile, \textit{FOL} not only optimizes the distance between different protected attributes but also enhances the dispersion among them, guiding the model towards greater fairness and effectiveness.
% Meanwhile, in addition to optimize distance between different protect attribute, FOL optimizes dispersion between different protect attribute to guide model to be more fair and effective.
Extensive experiments demonstrate the superiority of Fair-MoE. Detailed ablation studies provide evidence of the effectiveness of each component within Fair-MoE.

\bibliographystyle{named}
\bibliography{string}

\end{document}